\documentclass[journal,onecolumn]{IEEEtran}

\usepackage{graphicx}
\usepackage{amssymb}
\usepackage{subcaption}
\usepackage{siunitx}
\usepackage{url}
\usepackage{cite}
\usepackage{color}
\usepackage{xcolor}

\usepackage{prettyref}
\newrefformat{fig}{Figure~\ref{#1}}
\newrefformat{sec}{Section~\ref{#1}}
\newrefformat{tab}{Table~\ref{#1}}
\newrefformat{alg}{Algorithm~\ref{#1}}
\newrefformat{eq}{Equation~\ref{#1}}

\usepackage{hyperref}
\hypersetup{
    colorlinks,
    linkcolor={blue!80!black},
    citecolor={blue!80!black},
    urlcolor={blue!100!black}
}

\begin{document}

\title{Robust Shape Estimation for \\3D Deformable Object Manipulation}
\author{Tao Han, Xuan Zhao, Peigen Sun and Jia Pan
\thanks{The authors are with the Department of Mechanical and Biomedical Engineering, the City University of Hong Kong. This work was supported by HKSAR Research  Grants  Council  (RGC)  General  Research Fund (GRF) CityU 21203216, and NSFC/RGC Joint Research Scheme (CityU103/16-NSFC61631166002).}}

\maketitle

\begin{abstract}
Existing shape estimation methods for deformable object manipulation suffer from the drawbacks of being off-line, model dependent, noise-sensitive or occlusion-sensitive, and thus are not appropriate for manipulation tasks requiring high precision. In this paper, we present a real-time shape estimation approach for autonomous robotic manipulation of 3D deformable objects. Our method fulfills all the requirements necessary for the high-quality deformable object manipulation in terms of being real-time, model-free and robust to noise and occlusion. These advantages are accomplished using a joint tracking and reconstruction framework, in which we track the object deformation by aligning a reference shape model with the stream input from the RGB-D camera, and simultaneously upgrade the reference shape model according to the newly captured RGB-D data. We have evaluated the quality and robustness of our real-time shape estimation pipeline on a set of deformable manipulation tasks implemented on physical robots. Videos are available at \url{https://lifeisfantastic.github.io/DeformShapeEst/}
\end{abstract}

% section 1: introduction
\section{Introduction}
\label{sec:introduction}
\begin{figure}[t]
 \centering
 \includegraphics[width=0.8\linewidth]{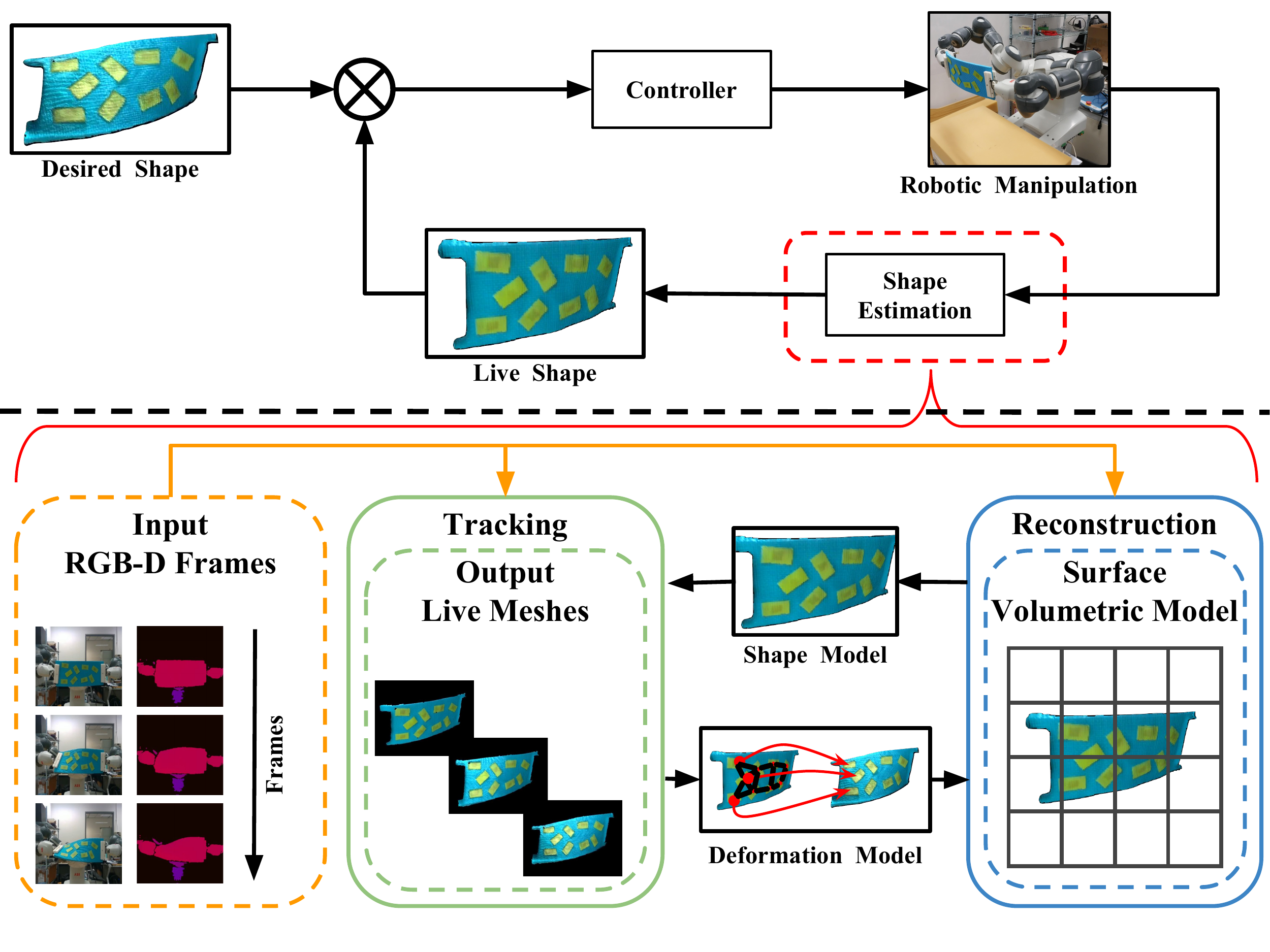}
 \caption{Overview of our shape estimation pipeline for deformable object manipulation. \textbf{Top row} is the entire control loop for deformable object manipulation. In this work, we focus on solving the shape estimation problem, as highlighted in the red-dashed box. An ideal shape estimation approach should fulfill the requirements of being real-time, model-free and robust to noise and occlusion. \textbf{Bottom row} illustrates the pipeline of our shape estimation system. Our system takes the frame sequence captured by a RGB-D camera as input. It is composed of two parallel threads, namely, \textit{tracking} and \textit{reconstruction}. These two threads estimate and update the deformation model and the shape model iteratively when new RGB-D frames are streaming into the system. Please refer to \prettyref{sec:overview} for more details about our pipeline.}
\label{fig:overview}
\end{figure}

Autonomous manipulation of deformable objects is an important and challenging topic in robotics, and recently it attracts much interest due to its potential applications in robot-assisted surgery~\cite{patil2010toward,Leizea:2017:RTV,navarro2013model,Yip:2014:MFC} and service robots, including garments folding~\cite{tanaka2018emd,doumanoglou2016folding,liu2018caging}, ironing~\cite{Li:2016:MSS}, and robot-assisted dressing~\cite{erickson2017does}. Despite the difference in their technical details employed for specific tasks, most existing systems for deformable object manipulation can be described using the same shape control framework as shown in the top row of \prettyref{fig:overview}. Given a desired shape of the target object, the robot system iteratively estimates the object's shape state according to either the sensor measurements or the simulation results, and applies the difference between the object's current shape state and the desired shape as an error signal to generate a control output to reduce or eliminate the error by deforming the object appropriately. The entire process repeats until the shape state converges to the desired value.

In this work, we focus on the shape estimation problem arising from the aforementioned control loop (as shown in the bottom row of \prettyref{fig:overview}). In particular, to represent the shape state of a deforming object, two models are needed: a \textit{shape model} encoding the geometry and texture of the deformable object's surface, and a \textit{deformation model} describing the deformation kinematics or dynamics of the object surface. When provided with sufficient prior knowledge about these two models, modern physically-based simulators such as~\cite{Bullet,Havok,PhysX} can provide a long-horizon prediction about the shape state of the underlying deformable object. As a result, recent work such as~\cite{schulman2013tracking,patil2010toward, sinha2014modeling,erickson2017does} embedded a physically-based engine into their pipeline and designed or learned a manipulation control according to the shape feedbacks provided by the simulator. The resulting model-based control could be robust to noise and occlusion, if the simulator has been carefully calibrated to be consistent with the real-world physics, which unfortunately is difficult in practice. In particular, the quality of the simulation is extremely sensitive to the model parameters, and this is considered as one main bottleneck of the model-based control for tasks involving deformable objects. In addition, running a physically-based simulation is time-consuming and thus infeasible for estimating fast or large deformation in real time.

For designing real-time and model-free manipulation system, recent work~\cite{navarro2016automatic,navarro2018fourier,hu20173d,jia2017manipulating} used the \textit{shape servoing} technique deriving from the traditional \textit{visual servoing} framework~\cite{hutchinson1996tutorial}. Instead of representing the object's shape via dense mesh structure as in the model-based method, the shape servoing method approximates the shape state using sparse key features extracted from image data. Because the feature descriptor is usually low-dimensional, the shape servoing method can learn the control policy between the shape feature and the manipulator motion directly in a data-driven manner. Such online policy learning framework makes the shape servoing method independent from an explicit deformation model to achieve shape control. However, existing methods in this direction still suffer from several drawbacks:

\begin{itemize}
\item{\textbf{Low-resolution shape modeling}}: Using sparse features as the shape feedback will omit some geometric details of the object. In other words, even when the feature representation of the object's current configuration perfectly matches the target feature vector, there is no guarantee that the object's shape can completely fit into its target shape. This limitation can be problematic for manipulation tasks which require high-precision goal reaching. As a result, a rich representation for the object deformation state is desirable. 
\item{\textbf{Noise-sensitive feature extraction}}: Most existing shape servoing approaches extract shape features from a single frame of image. The extracted vector can be unreliable for closed-loop control due to the noises in the feedback image that are ubiquitous in real robotic systems. As a result, we need a sophisticated method to obtain robust features from a sequence of feedback inputs.
\item{\textbf{Occlusion-sensitive feature correspondences}}: Existing shape servoing methods rely on a feature descriptor to determine inter-frame feature correspondences. However, many deformable objects lack visually significant feature points, and thus they must have additional markers mounted on the surface to provide reliable feedbacks. 
Such marker-based workaround leads to inconvenience for practical applications. Moreover, most shape servoing systems assume that the full state of the object's surface can be observed all the time during the task. As a result, when some feature points or markers become invisible to the visual sensor due to occlusion, these systems may fail to capture enough feature vectors for servoing control.
\end{itemize}

The aforementioned problems in both model-based method and shape servoing method motivated us to propose a novel shape estimation method in this paper. Our method satisfies the requirements of being real-time, model-free and robust to noise and occlusion, and thus can be easily embedded into current robotic systems for autonomous manipulation of general 3D soft objects. In our work, we further divide the shape estimation problem into two subproblems, namely \textit{tracking} and \textit{reconstruction} (as shown in the bottom row of \prettyref{fig:overview}). In the tracking phase, we estimate an inter-frame deformation model through non-rigid registration between a reference shape model and the depth images provided by an RGB-D camera. In the reconstruction phase, we integrate multiple RGB-D images into the reference shape model according to the estimated deformation model. One key contribution of our work is that our simultaneous tracking and reconstruction framework can capture the surface model of a deforming object, while gradually complete and refine its details based on new RGB-D measurements. Because the generated surface model is of high precision and is also robust to single-frame noise and occlusion, it serves as an excellent feedback signal for shape control.

% section 2: related work
%\input{sec_2}

% section 3: preliminaries
\section{Overview}
\label{sec:approach}

\begin{table}
\label{tab:nomenclature}
\begin{tabular}{p{.03\textwidth}p{.05\textwidth}p{.8\textwidth}}
$\mathcal{F}^{0},\mathcal{F}^{t}$    & & reference frame, live frame \\
$\mathcal{M}^{0},\mathcal{M}^{t}$    & & mesh model defined in the corresponding frame space \\
$\mathbf{x}^{0},\mathbf{x}^{t}$      & & voxel defined in the corresponding frame space \\
$\mathbf{v}^{0}_m,\mathbf{n}^{0}_m$          & & the $m$-th vertex element and normal element of mesh $\mathcal{M}^{0}$ \\
$\mathbf{v}^{t}_m,\mathbf{t}^{t}_m$          & & the $m$-th vertex element and normal element of mesh $\mathcal{M}^{t}$ \\
$\mathbb{V}$                         & & reference volume defined in the space of $\mathcal{F}^{0}$ \\
$\mathcal{D}(\cdot)$                 & & TSDF component of the voxel in $\mathbb{V}$ \\
$\mathcal{C}(\cdot)$                 & & color component of the voxel in $\mathbb{V}$ \\
$\Omega(\cdot)$                      & & weight component of the voxel in $\mathbb{V}$ \\
$d(\cdot)$                           & & TSDF component of the voxel contributed by $\mathcal{F}^{t}$ \\
$c(\cdot)$                           & & color component of the voxel contributed by $\mathcal{F}^{t}$ \\
$\omega(\cdot)$                      & & weight component of the voxel contributed by $\mathcal{F}^{t}$ \\
$\mathcal{G}$                        & & deformation model of the target soft object \\
$\mathbf{T}\textsubscript{rigid}$    & & rigid component separated from $\mathcal{G}$ \\
$\mathcal{G}\textsubscript{graph}$   & & non-rigid component of $\mathcal{G}$ represented as graph \\
$\mathbf{g}_{i}$                     & & the $i$-th node of the graph $\mathcal{G}\textsubscript{graph}$ defined in the space of $\mathcal{F}^{0}$ \\
$\mathbf{T}_{i}$                     & & local deformation defined in $\mathbf{g}_{i}$ \\
$\sigma_{i}$                         & & effective radius of $\mathbf{T}_{i}$ \\
$\mathcal{N}_{i}$                    & & neighbor set of $\mathbf{g}_{i}$ \\
$\mathcal{W}(\cdot;\mathcal{G})$     & & deformation function maps $\mathcal{M}^{0}$ to $\mathcal{M}^{t}$, parameterized by $\mathcal{G}$ \\
$C^{t}(\cdot)$                       & & color map of $\mathcal{F}^{t}$ \\
$D^{t}(\cdot)$                       & & depth map of $\mathcal{F}^{t}$ \\
$V^{t}(\cdot)$                       & & vertex map extracted from $D^{t}(\cdot)$ \\
$N^{t}(\cdot)$                       & & normal map extracted from $D^{t}(\cdot)$ \\
\end{tabular}
\caption{Nomenclature}
\end{table}

In this section, we first present the mathematical notation to define the shape and deformation models employed in our work. Then we outline the pipeline of our simultaneous shape tracking and reconstruction framework.
\begin{figure}[t]
 \centering
 \includegraphics[width=0.8\linewidth]{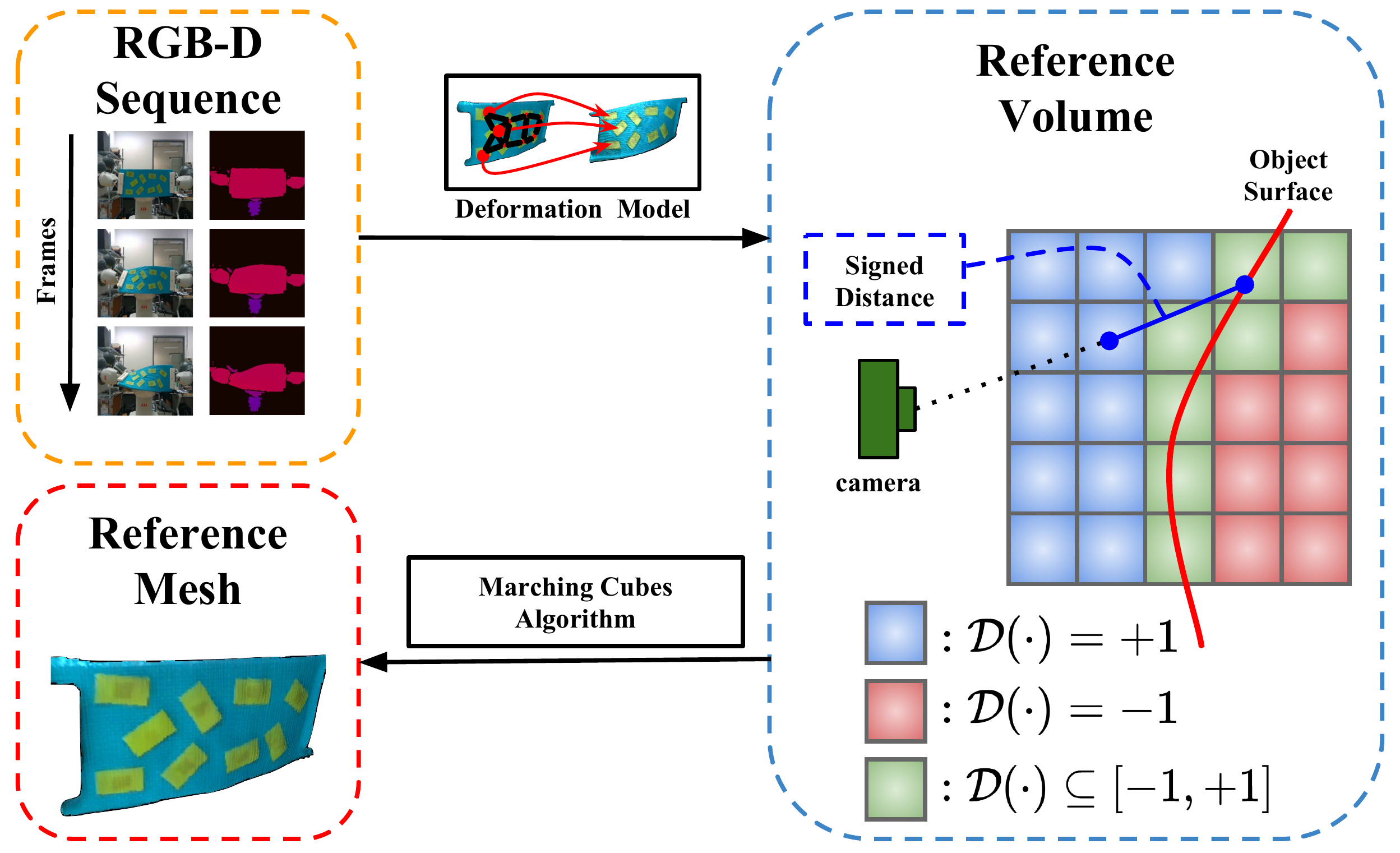}
 \caption{Illustration of the reference volume $\mathbb{V}$ in 2D case. The signed distance value of each voxel is defined as the distance from its center to its closest point on the surface (shown as the solid blue, positive when the voxel is on the ``front'' side of the surface and negative when it is on the ``back'' side of the surface). We further truncate the signed distance with a threshold to make its value belong to the range $[-1, +1]$. A reference mesh can be extracted from the reference volume $\mathbb{V}$ by locating the zero-level of the TSDF value.}
 \label{fig:tsdf}
\end{figure}

\subsection{Shape Representation}
\label{sec:shape_rep}
The main objective of deformable object manipulation is to deform the object's surface from an initial shape into a desired shape. The inner state of the target object is actually ignored for shape control in most current methods. Therefore in this work we are only interested in how to use an appropriate shape representation to model the surface state. To generate high-quality shape model containing rich geometry and texture details, one commonly used solution, similar to the model-based method, is to represent the surface based on the mesh structure extracted from the RGB-D image data~\cite{pan2017superpixels}. While such mesh-based representation is suitable to serve as the reference model for shape tracking, its graph structure makes it hard to be fused with multiple image frames for shape reconstruction. Instead, encoding the surface geometry and texture into a 3D volumetric grid structure is more feasible since the shape reconstruction progress can be implemented efficiently via parallel operation on the grids.

To combine the advantages of both representations, we follow previous work~\cite{newcombe2011kinectfusion, newcombe2015dynamicfusion,cao2017extracting} by projecting multiple image frames data back into the space of a reference frame $\mathcal{F}^{0}$ (which is usually set as the initial frame) according to the estimated inter-frame deformations and then integrating these frames into a \textit{reference mesh} $\mathcal{M}^{0}$. For efficient image integration, the reference mesh $\mathcal{M}^{0}$ is maintained via a discrete truncated signed distance function (TSDF) volume (as illustrated in \prettyref{fig:tsdf}), which we denote as the \textit{reference volume} $\mathbb{V}$. In this reference volume $\mathbb{V}$, the surface geometry is voxelized as $\{ \mathcal{D}(\mathbf{x}^{0}), \Omega(\mathbf{x}^{0}) \}$, where $\mathcal{D}(\mathbf{x}^{0}) \subseteq [-1, +1]$ encodes the truncated signed distance value for each voxel $\mathbf{x}^{0}$, and $\Omega(\mathbf{x}^{0}) \subseteq [0, 1]$ is the associated weight. The content of each voxel will be updated independently for image integration. To obtain a high-quality mesh with texture, we also maintain the RGB information $\mathcal{C}(\mathbf{x}^{0}) \subseteq [0,255]^{3}$ in each voxel. As a summary, our reference volume can be represented as $\mathbb{V}(\mathbf{x}^{0}) = \{\mathcal{D}(\mathbf{x}^{0}), \mathcal{C}(\mathbf{x}^{0}), \Omega(\mathbf{x}^{0})\}$.

From the reference volume $\mathbb{V}$, we extract the reference mesh $\mathcal{M}^{0}$ with Marching Cubes algorithm ~\cite{lorensen1987marching}. The reference mesh $\mathcal{M}^{0}$ can be further deformed according to the estimated inter-frame deformation model to obtain the \textit{live mesh} $\mathcal{M}^{t}$, which indicates the object shape in the live frame $\mathcal{F}^{t}$. For the convenience of discussion, we define the mesh vertices and corresponding normals as $\{ \mathbf{v}_{m}, \mathbf{n}_{m} \}_{m=1}^{M}$ in this work, where $M$ is the number of vertices in the mesh.

\subsection{Deformation Representation}
\label{sec:deform_rep}
To achieve simultaneous shape tracking and reconstruction, we need a model to formulate the deformation from the reference frame $\mathcal{F}^{0}$ to the live frame $\mathcal{F}^{t}$. In consideration of our model-free requirement, skeleton-based kinematic models for articulated objects are infeasible to represent the deformation of general soft objects. One possible solution is to model the deformation based on Eulerian (grid-based) method or Lagrangian (particle-based) method used in fluid simulation~\cite{bridson2015fluid}. While both methods can provide high-quality representation for general deformation based on their dense structures, such high-dimensional models are not feasible for real-time estimation.

As a trade-off between complexity and precision, we employ the sparse deformation graph model~\cite{sumner2007embedded} with reduced dimensions for real-time implementation. In this method, the graph nodes are uniformly sampled from the mesh model to have a layout roughly conform to the object's shape. Then the whole deformation is divided into a set of local transformations, which are then assigned to the graph nodes one-to-one. The graph nodes have overlapping domain of influence with their neighboring nodes in local transformations. Thus for any given point in the nearby space of the graph nodes, a smooth deformation function can be computed via interpolation of the local transformations in the point's nearest graph nodes.

Moreover, to avoid over-fitting of the deformation graph model during estimation, a regularization constraint is also needed. In our work, we regularize the deformation graph via the widely used \textit{as-rigid-as-possible} (ARAP) constraint~\cite{sorkine2007rigid}, which penalizes inconsistent local transformations between neighboring graph nodes. Such penalty function is usually be represented geometrically as the graph edges.

Except the graph model, we further divide the global rigid component from the total deformation and model it separately. Overall, our deformation model can be represented as $\mathcal{G} = \mathbf{T}_{\text{rigid}} \cup \mathcal{G}\textsubscript{graph}$. Here $\mathbf{T}_{\text{rigid}} \in \text{SE}(3)$ defines the separated global rigid transformation. $\mathcal{G}\textsubscript{graph}$ denotes the non-rigid component of $\mathcal{G}$ represented in the graph model. We further parameterize the graph model as $\mathcal{G}\textsubscript{graph} = \{ \mathbf{T}_{i}, \mathbf{g}_{i}, \sigma_{i}, \mathcal{N}_{i} \}_{i=1}^{K}$. $\mathbf{T}_{i} \in \text{SE}(3)$ indicates the local transformation in the $i$-th node. $\mathbf{g}_{i}$ represents the position of the $i$-th node in the reference frame $\mathcal{F}^{0}$. $\sigma_{i}$ defines the effective radius of $\mathbf{T}_{i}$. The neighbor set $\mathcal{N}_{i}$ contains the indices of those nodes which are connected with the $i$-th node by graph edges. These nodes $\{ \mathbf{g}_{j} \vert j \in \mathcal{N}_{i} \}$ are considered as the closest neighbors of the $i$-th node. Note that in our method, $\{ \mathbf{g}_{i}, \sigma_{i}, \mathcal{N}_{i} \}_{i=1}^{K}$ remains constant during estimation, and thus our deformation model can be fully parameterized as $\mathcal{G} = \{ \mathbf{T}_{\text{rigid}} \} \cup \{ \mathbf{T}_{i} \}_{i=1}^{K}$.

To deform the reference mesh $\mathcal{M}^{0}$ according to the above model $\mathcal{G}$, we first assign each mesh vertex $\mathbf{v}_{m}^{0} \in \mathcal{M}^{0}$ to its $k$-nearest nodes $\mathcal{S}_{m} \subseteq \{ 1, \ldots, K \}$ on the graph model of $\mathcal{G}$ based on a set of skinning weights $\{ \omega_{m,k} : k \in \mathcal{S}_{m} \} \subseteq [0,1]$. The skinning weight is calculated as $\omega_{m,k} = \frac{1}{Z} \exp ( - \Vert \mathbf{v}_{m}^{0} - \mathbf{g}_{k} \Vert^{2} / 2\sigma_{k}^{2} )$, where $Z$ is a normalization factor ensuring $\sum_{k} \omega_{m,k} = 1$. Then we calculate the deformed vertex $\mathbf{v}_{m}^{t}$ in the live frame $\mathcal{F}^{t}$ based on the following blending function:
\begin{equation}
\label{eq:blending func}
 \mathbf{v}_{m}^{t} = \mathcal{W}(\mathbf{v}_{m}^{0}; \mathcal{G}) = \mathbf{T}_{\text{rigid}} \sum_{k \in \mathcal{S}_{m}} \omega_{m,k} \mathbf{T}_{k} \mathbf{v}_{m}^{0}.
\end{equation}
Similarly, the corresponding normal $\mathbf{n}_{m}^{0}$ of the vertex $\mathbf{v}_{m}^{0}$ can be deformed using the following blending function:
\begin{equation}
 \mathbf{n}_{m}^{t} = \mathcal{W}(\mathbf{n}_{m}^{0}; \mathcal{G}) = \mathbf{T}_{\text{rigid}} \sum_{k \in \mathcal{S}_{m}} \omega_{m,k} \mathbf{T}_{k} \mathbf{n}_{m}^{0}.
\end{equation}

\subsection{Our System}
\label{sec:overview}

As demonstrated in \prettyref{fig:overview}, our system takes the image stream provided by an RGB-D camera as input. It is composed of two parallel threads: tracking and reconstruction. The tracking thread takes charge of the real-time estimation of the deformation model $\mathcal{G}$. It aligns the live RGB-D frame $\mathcal{F}^{t}$ with the reference mesh model $\mathcal{M}^{0}$ for geometric consistency. To capture the surface geometry for alignment, the tracking thread extracts dense features from the received depth map $D^{t}$ of the live frame $\mathcal{F}^{t}$, including a vertex map $V^{t}$ and a corresponding normal map $N^{t}$. At the core of this thread is a highly-efficient GPU solver which optimizes the deformation model $\mathcal{G}$ for frame-to-model alignment under the regularization constraint. We implement the optimization solver based on a kernel merged \textit{preconditioned conjugate gradient} (PCG) algorithm using CUDA.

Given the estimation of the deformation model $\mathcal{G}$, the reconstruction thread computes the voxel-to-pixel correspondences between the reference volume $\mathbb{V}$ and the live frame $\mathcal{F}^{t}$ and updates the content in each voxel of $\mathbb{V}$ accordingly. After the previous volume fusion operation, the reconstruction thread extracts a new reference mesh $\mathcal{M}^{0}$ from the reference volume $\mathbb{V}$ and obtains the associated live mesh $\mathcal{M}^{t}$ based on the estimated deformation model $\mathcal{M}^{t} = \mathcal{W}(\mathcal{M}^{0}; \mathcal{G})$.

% section 4: tracking
\subsubsection{Tracking}
\label{sec:tracking}
The objective of the tracking thread is to provide accurate estimation of the deformation model to assist the reconstruction thread. In the tracking step, we optimize the deformation model to obtain the best frame-to-model alignment between the live frame $\mathcal{F}^{t}$ and the reference mesh model $\mathcal{M}^{0}$.

To estimate the deformation model $\mathcal{G}$, we formulate the following energy function $E(\mathcal{G})$:
\begin{equation}
\label{eq:energy func}
E(\mathcal{G}) = \lambda_{\text{data}}E_{\text{data}}(\mathcal{G}) + \lambda_{\text{reg}}E_{\text{reg}}(\mathcal{G}),
\end{equation}
where $E_{\text{data}}(\mathcal{G})$ is the data term which penalizes the misalignment between the reference mesh model $\mathcal{M}^{0}$ and the dense features extracted from the live image frame $\mathcal{F}^{t}$. $E_{\text{reg}}(\mathcal{G})$ is the regularization term which penalizes the inconsistent local transformations between neighboring nodes in the graph model of $\mathcal{G}$. $\lambda_{\text{data}}$ and $\lambda_{\text{reg}}$ denote the associated weights of these two terms.

\noindent{\textbf{Data Term}}
To measure the misalignment between the reference mesh model $\mathcal{M}^{0}$ and the live frame $\mathcal{F}^{t}$, we first calculate a vertex map $V^{t}$ and a normal map $N^{t}$ from the depth map $D^{t}$ to represent the geometric features of the live frame $\mathcal{F}^{t}$. Then we deform the reference mesh vertices $\mathbf{v}^{0}$ and normals $\mathbf{n}^{0}$ according to the estimation of the deformation model $\mathcal{G}$ to obtain their prediction represented in the live frame $\mathcal{F}^{t}$. In particular, the predicated vertices are $\mathbf{\tilde{v}}^{t}(\mathcal{G}) = \mathcal{W}(\mathbf{v}^{0}; \mathcal{G})$ and the predicated normals are $\mathbf{\tilde{n}}^{t}(\mathcal{G}) = \mathcal{W}(\mathbf{n}^{0}; \mathcal{G})$. Finally, we quantify the misalignment between the predicted vertices $\mathbf{\tilde{v}}^{t}$ and the geometric features $\{ V^{t}, N^{t} \}$ based on the point-to-plane error function widely used in the Iterative Closest Point (ICP) algorithm. As a result, we represent the geometric data term as
% \begin{equation}
% E_{\text{geo}}(\mathcal{G}) = \sum_{m \in \mathcal{V}_{m}}\Vert \mathbf{n}_{d}^{\top} (\mathbf{\tilde{v}}_{m}^{t}(\mathcal{G}) - \mathbf{v}_{d}) \Vert^{2}_{2},
% \end{equation}
\begin{equation}
\label{eq:geo_data_term}
E_{\text{geo}}(\mathcal{G}) = \sum_{m}\Vert \mathbf{n}_{d}^{\top} (\mathbf{\tilde{v}}_{m}^{t}(\mathcal{G}) - \mathbf{v}_{d}) \Vert^{2}_{2},
\end{equation}
where $\{ \mathbf{v}_{d}, \mathbf{n}_{d} \}$ denote the 3D-to-2D projected correspondences of the predicted vertex $\mathbf{\tilde{v}}_{m}^{t}$ in the feature map $\{ V^{t}, N^{t} \}$. 

\noindent{\textbf{Regularization Term}}
The deformation graph model can easily become over-fitting if it is not well regularized during estimation. To solve this problem, one widely used template-free method for general soft objects is to introduce the ARAP constraint. In our work, we encode the neighboring relationship of the ARAP constraint into the deformation graph neighbor set $\{ \mathcal{N}_{i} \}_{i=1}^{K}$. Based on the neighbor set, we define the regularization term as
\begin{equation}
E_{\text{reg}} = \sum_{i=1}^{K} \sum_{j \in \mathcal{N}_{i}} \Vert \mathbf{T}_{i}\mathbf{g}_{i}^{t} - \mathbf{T}_{j}\mathbf{g}_{i}^{t}  \Vert_{2}^{2}.
\end{equation}

% section 5: reconstruction
\subsubsection{Reconstruction}
\label{sec:reconstruction}

The reconstruction thread takes the live frame $\mathcal{F}^{t}$ and the newly estimated deformation model $\mathcal{G}$ as input. It updates the surface geometry and texture by integrating multiple image frames incrementally into the reference volume $\mathbb{V}$. Because the surface geometry and texture are encoded in the volumetric structure, we refer this procedure as volume fusion. 

We implement the volume fusion operation based on a non-rigid projective fusion approach \cite{newcombe2011kinectfusion}. In this approach, we first scan the voxels of $\mathbb{V}$ and get their positions in the reference frame space, which are denoted as $\mathbf{x}^{0}$. Then we calculate the corresponding deformed positions $\mathbf{x}^{t} = \mathcal{W}^{0,t}(\mathbf{x}^{0}; \mathbb{G})$ in the live frame based on the blending function in \prettyref{eq:blending func}. The deformed voxels $\mathbf{x}^{t}$ are projected into the live frame image map to find their corresponding pixels $\mathbf{u}^{t}$. Thus we determine the voxel-to-pixel correspondence $\{ \mathbf{x}^{0} \leftrightarrow \mathbf{u}^{t} \}$ between the reference volume $\mathbb{V}$ and the live frame image $\mathcal{F}^{t}$. For each voxel $\mathbf{x}^{0}$, we calculate its new TSDF component $d(\mathbf{x}^{0})$ and color component $c(\mathbf{x}^{0})$ contributed by the live frame $\mathcal{F}^{t}$ as
\begin{equation}
d(\mathbf{x}^{0}) = \max \left( \min \left(\frac{D^{t}(\mathbf{u}^{t}) - \lfloor \mathbf{x}^{t} \rfloor_{z}}{\tau}, 1 \right), -1 \right),
\end{equation}
and
\begin{equation}
c(\mathbf{x}^{0}) = C^{t}(\mathbf{u}^{t}),
\end{equation}
respectively. Here $D^{t}(\cdot)$ and $C^{t}(\cdot)$ denote the depth image and color image of the live frame $\mathcal{F}^{t}$. $\lfloor \mathbf{x} \rfloor_{z}$ represents the position of point $\mathbf{x}$ along Z-axis. $\tau$ is the truncated threshold of TSDF value. Besides, we assign a weight $\omega(\mathbf{x}^{0})$ to the new components. Finally, we update the reference volume $\mathbb{V}$ as
\begin{equation}\label{eq:tsdf_fusion}
\mathcal{D}(\mathbf{x}^{0}) \leftarrow \frac{\mathcal{D}(\mathbf{x}^{0})\Omega(\mathbf{x}^{0}) + d(\mathbf{x}^{0})\omega(\mathbf{x}^{0})}{\Omega(\mathbf{x}^{0}) + \omega(\mathbf{x}^{0})},
\end{equation}
\begin{equation}\label{eq:color_fusion}
\mathcal{C}(\mathbf{x}^{0}) \leftarrow c(\mathbf{x}^{0}),
\end{equation}
\begin{equation}
\Omega(\mathbf{x}^{0}) \leftarrow \min \left( \Omega(\mathbf{x}^{0}) + \omega(\mathbf{x}^{0}),\Omega_{\max} \right),
\end{equation}
where $\Omega_{\text{max}}$ is the upper threshold of the weight. In \prettyref{eq:color_fusion}, we do not update the color component via integration as in \prettyref{eq:tsdf_fusion}. The main reason here is that our current work does not model and track the light environment and material albedo. As a result, setting the color component $\mathcal{C}(\mathbf{x}^{0})$ directly as its new value $c(\mathbf{x}^{0})$ in live frame can obtain better response than data fusion.

% section 6: experiments
\section{Experiments and Results}
\label{sec:exps_and_results}
\begin{figure}[!htb]
 \centering
 \includegraphics[width=0.9\linewidth]{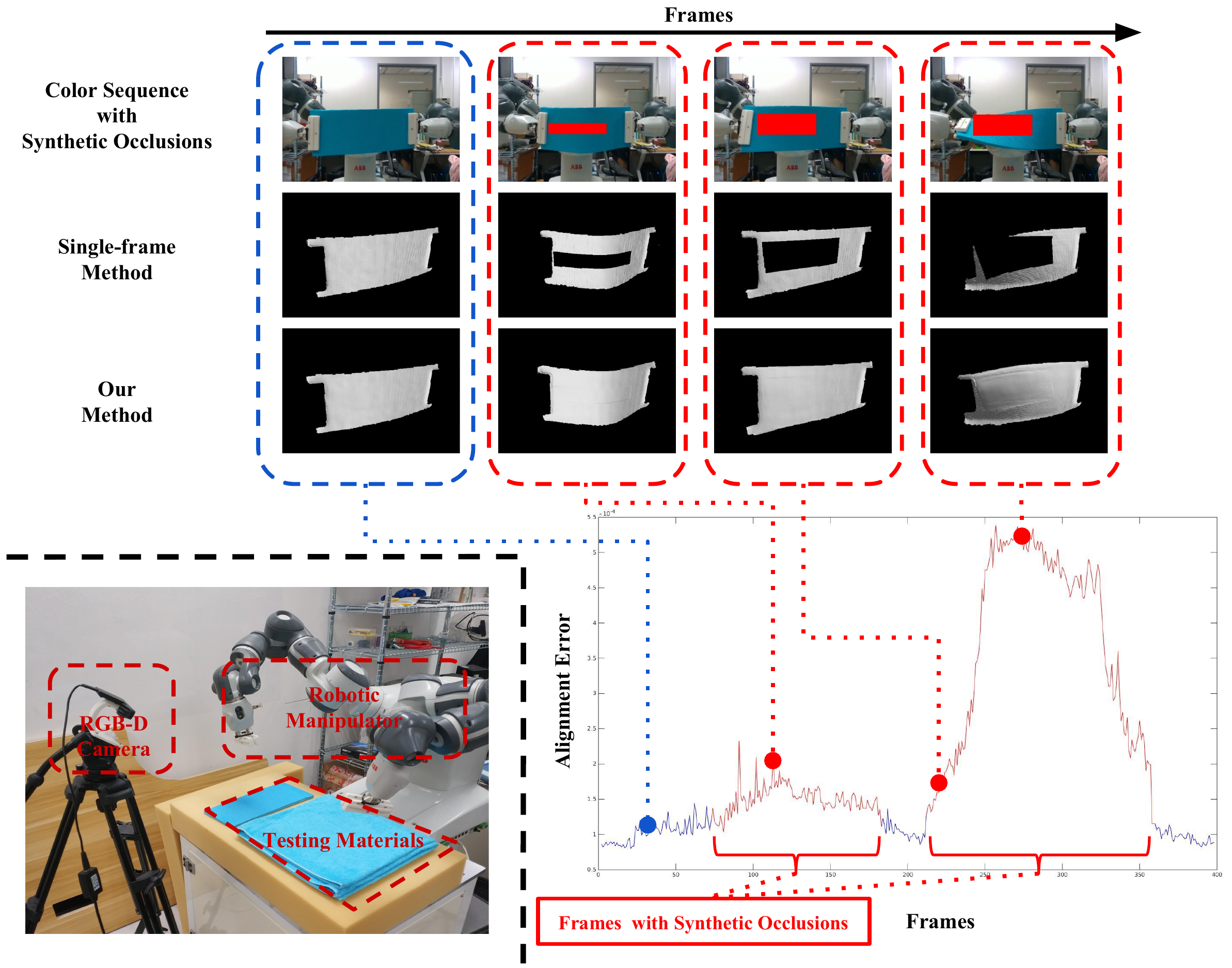}
 \caption{\textbf{Bottom left corner} illustrates our experiment setup for a typical deformable object manipulation tasks using a dual arm robot. \textbf{Top part} shows the results of the plastic sheet bending task containing synthetic occlusions. The synthetic occlusion masks are highlighted as the red-solid boxes in the first row. \textbf{Bottom part} shows the alignment errors between the generated mesh models of our method and the raw image sequence without synthetic occlusions. Because our method approximates the deformation behavior of the target object based on the ARAP constraint, it can provides shape estimation even for the unobserved parts of the surface.}
 \label{fig:exp_setup}
\end{figure}

We implement our shape estimation pipeline on a desktop PC with Intel Core i7 3.4GHz CPU, 32GB of RAM and an NVIDIA GeForce GTX 1080 GPU. To setup the working environment of typical deformable object manipulation tasks, we employ one dual-arm robot (ABB YuMi, with seven degrees-of-freedom in each arm) to perform demonstrations with different materials. Besides, we take the RGB-D data provided by an Intel RealSense SR300 camera as input. The entire experimental setup is shown in the bottom left corner of \prettyref{fig:exp_setup}.

Because we encode the reference surface model $\mathcal{M}^{0}$ explicitly into the volumetric structure $\mathbb{V}$, the real-time performance of our pipeline largely depends on the parameters of the volumetric model $\mathbb{V}$, including the volume's dimension in voxels, the truncated threshold of TSDF value $\tau$, the weight of the newly captured TSDF component $\omega$, and the upper-bound weight of reference TSDF component $\Omega_{\max}$. In our experiment, we pre-defined a \SI{0.7}{m^3} cubic space as the reference volume $\mathbb{V}$ and discretized it based on $512^{3}$ voxels. Thus the actual resolution of the volumetric model is $731^{3}$ voxels per cubic meters. Besides, we set $\tau$, $\omega(\cdot)$ and $\Omega_{\max}$ as constants, in particular, $\tau = 0.01m$, $\omega(\cdot) = 1$, and $\Omega_{\max} = 32$. We measured the runtime cost of each main computational components in our pipeline during our experiment, including \SI{3}{ms} for preprocessing (e.g., depth image filtering, vertex map and normal map extracting), \SI{35}{ms} for deformation tracking, and \SI{10}{ms} for volume fusion. On average, our pipeline runs at \SI{50}{ms} per frame to track and reconstruct the surface of all deforming targets employed in our experiment, which satisfies the real-time requirement of most robotic applications.

As previously mentioned, one main advantage of our method is that it is robust to occlusion. Such robustness is crucial for applications involving human-machine cooperation, where the human body may occlude the target object from the camera. To demonstrate the advantage of our method, we test it with a plastic sheet bending task. In this task, we let the robotic arm deform the target sheet in front of an RGB-D camera. During the task, we introduce some synthetic occlusions into the captured RGB-D stream and make the corresponding surface areas become invisible to our system. The synthetic occlusion masks are illustrated as the red-solid boxes in the top row of \prettyref{fig:exp_setup}. To compared with most previous work~\cite{navarro2016automatic,navarro2018fourier,hu20173d,jia2017manipulating} which employed single-frame data for shape estimation, we show the corresponding surface mesh model extracted from each frame in the second row of \prettyref{fig:exp_setup}. Note since we encoded all surface information captured by the single-frame image into such mesh model, it indicates the input data adopted by the aforemetioned work. In our experiment, we extracted such mesh model by projecting the recently captured singe-frame data into a new TSDF volume, and then locating the zero-level surface based on the Marching Cubes algorithm~\cite{lorensen1987marching} without consideration of the previously observed data. We refer to such method as the \textit{single-frame method}. The mesh models generated by our method are illustrated in the third row of \prettyref{fig:exp_setup}. As we can observe, the single-frame method cannot capture the geometry of the occluded part of the object. Our method, on the other hand, approximates the deformation behavior of the unobserved part based on the ARAP constraint and generates a complete shape estimation accordingly.

\begin{figure}[t]
 \centering
 \includegraphics[width=0.9\linewidth]{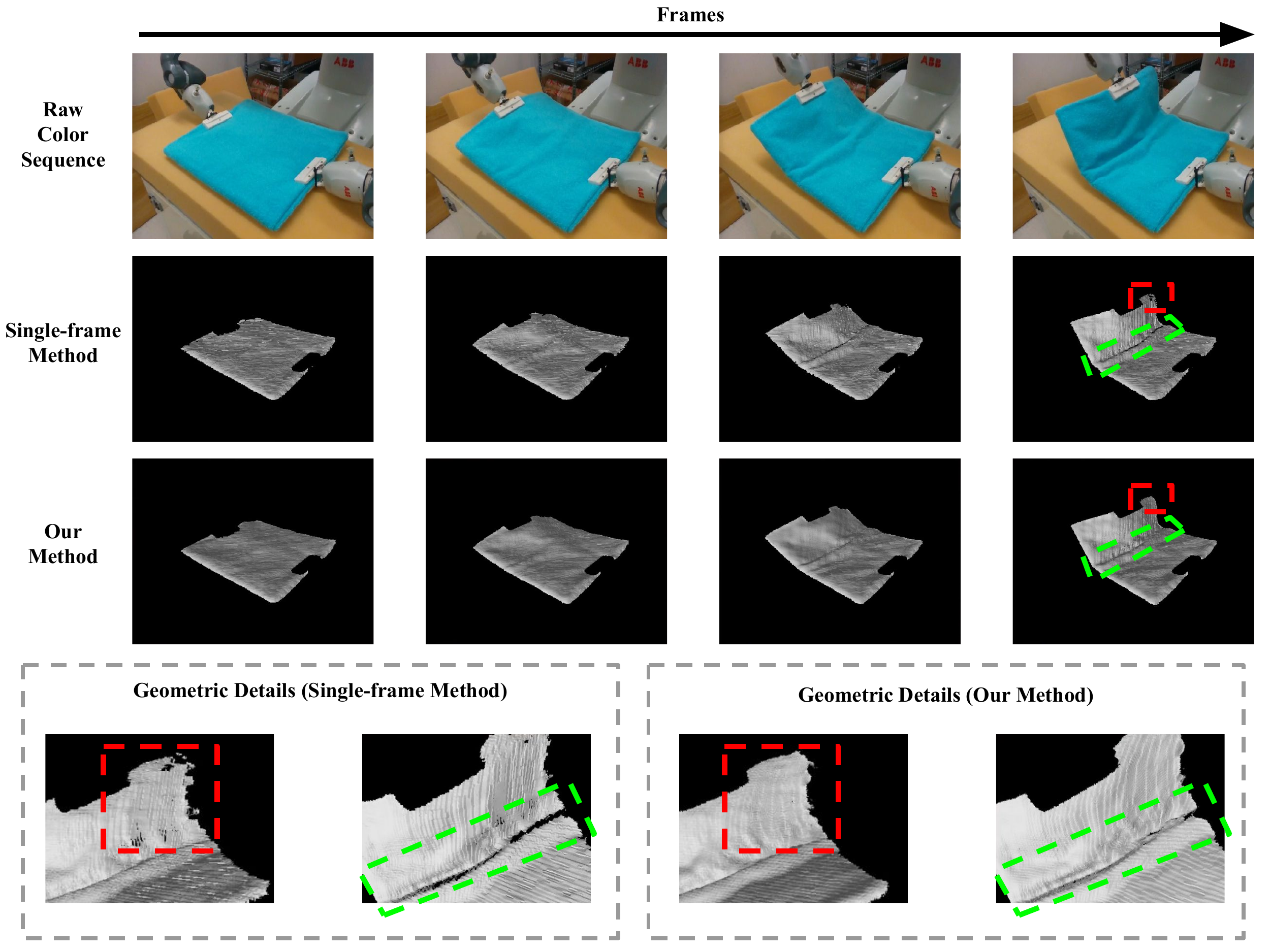}
 \caption{Shape estimation results in our simulated towel folding task. \textbf{Top row} shows the raw color sequence captured by a RGB-D camera. \textbf{Middle row} shows the live mesh model reconstructed from the single-frame depth data. \textbf{Bottom row} shows the live mesh model generated by our method. We zoom in on the highlighted areas in the last frame to demonstrate their geometric details.}
 \label{fig:exp_noise}
\end{figure}

Because the occlusions in the aforementioned experiment are synthetic, we further evaluated the accuracy of our method, especially for the estimate of the occluded surface part, by measuring the non-rigid alignment error between the reconstructed mesh model and the raw image data without synthetic occlusions based on the data term in \prettyref{eq:geo_data_term}. We plot the alignment errors measured at different time instants in the bottom right corner of \prettyref{fig:exp_setup}. Consider the occluded surface part is totally driven by the ARAP constraint in our method, there is no doubt a gap between our deformation graph model and the object's real deformation behavior. When the occluded part undergoes small deformation as in the second column of \prettyref{fig:exp_setup}, the ARAP constraint holds well and the alignment error is quite close to the case without occlusion. The mentioned gap will become obvious when the occluded part undergoes large deformation as in the fourth column of \prettyref{fig:exp_setup}. However, even in the latter case, the ARAP constraint can still contribute to a compelling mesh reconstruction. Moreover, its independence of object prior knowledge is essential for our model-free implementation.

To demonstrate the robustness of our method to sensor data noise, we design another folding towel task for testing. Again, we compare the reconstruction results of two different methods in \prettyref{fig:exp_noise}, including the single-frame method and our method. Because the RGB-D camera cannot provide stable depth measurement for the wrinkle areas (as highlighted in the green rectangle) and the nearly parallel areas (as highlighted in the red rectangle) on the folded towel, the single-frame reconstructions omit some important geometric details for the shape feedback. This problem exists in most current shape servoing methods. Instead, our method updates the geometry of the surface model via efficient image data fusion, and is capable of providing continuous and smooth mesh reconstruction.

% section 7: conclusion and discussion
\section{Conclusion, Limitations and Future Work}
We present a novel shape estimation method to provide reliable shape feedback for the deformable object manipulation problem. A series of experiments are conducted to show the advantages of our method in terms of being real-time, model-free and robust to noise and occlusion. All these features make our method promising to be embedded into current robotic manipulation systems for challenging applications.

Our method still has some limitations. First, our method relies on high-precision deformation estimation for consistent and accurate shape reconstruction. In other words, when the estimation step fails in some cases, the drift error will be accumulated into the reconstruction result and cannot be corrected. One possible solution to this problem is to add a module into the pipeline which does not rely on the deformation estimates provided by the tracking thread for drift correction.

Second, our deformation model, especially the ARAP regularization term, cannot always hold when the target object undergoes large deformations or complex topological changes. The reasons for such limitation are two-fold: On one hand, to stay within our computational budget for real-time application, we approximated the ARAP regularization term by penalizing inconsistent local deformations close in Euclidean space rather than on the mesh manifold, which unfortunately introduces a gap between the employed deformation model and the real-world physics. On the other hand, the proposed system lacks the ability for perceiving and inferring the surface topology. Thus even if the ARAP constraint is formulated strictly according to the distance on the mesh manifold, it is still difficult for the system to track and reconstruct the deforming surface undergoes fast or complex topological changes. Due to these reasons, we only present experiments based on deforming objects with simple topology and geometry. Note improving the system's robustness to handle topological changes is still challenging for all model-free methods in related fields, and we sincerely believe that a topological segmentation front-end is essential for solving such a problem.

Besides resolving above limitations, for our future work, we will also present a complete shape control pipeline embedded with our shape estimation method.

\bibliographystyle{./IEEEtran}
\bibliography{./cis_ref}

% Xuan Zhao
% \address{Department of Mechanical and Biomedical Engineering\\
% City University of Hong Kong\\
% Tat Chee Avenue, Kowloon, Hong Kong SAR\\
% \email{xuan.zhao@my.cityu.edu.hk}}

% Peigen Sun
% \address{Department of Mechanical and Biomedical Engineering\\
% City University of Hong Kong\\
% Tat Chee Avenue, Kowloon, Hong Kong SAR\\
% \email{}}

% Jia Pan
% \address{Department of Mechanical and Biomedical Engineering\\
% City University of Hong Kong\\
% Tat Chee Avenue, Kowloon, Hong Kong SAR\\
% \email{}}

% \received{September 26, 2014}\\
% \accepted{March 11, 2015}}
\end{document}